\documentclass[runningheads]{llncs}
\usepackage{graphicx}
\usepackage{amsmath,amssymb} 
\usepackage{color}
\usepackage{float}
\begin{document}
\pagestyle{headings}
\mainmatter

\title{An Acceleration Scheme for Memory Limited, Streaming PCA} 

\author{Salaheddin Alakkari and John Dingliana}
\institute{Graphics, Vision and Visualisation Group (GV2),\\
School of Computer Science and Statistics,\\
Trinity College Dublin}

\maketitle


\begin{abstract}
    In this paper, we propose an acceleration scheme for online memory-limited PCA methods. Our scheme converges to the first $k>1$ eigenvectors in a single data pass. We provide empirical convergence results of our scheme based on the spiked covariance model. Our scheme does not require any predefined parameters such as the eigengap and hence is well facilitated for streaming data scenarios. Furthermore, we apply our scheme to challenging time-varying systems where online PCA methods fail to converge. Specifically, we discuss a family of time-varying systems that are based on Molecular Dynamics simulations where batch PCA converges to the actual analytic solution of such systems.
\end{abstract}
		\begin{figure}[h]
		\centering
			\includegraphics[width=0.9\linewidth]{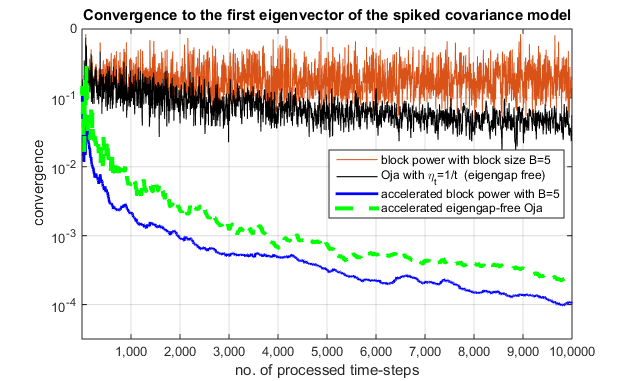}
		\caption{\label{fig:example}
			An overview example showing the convergence to the first eigenvector of the spiked covariance model before and after applying our acceleration scheme.}
	\end{figure}

\section{Introduction}
Principal Component Analysis (PCA) is one of the most important machine learning and dimensionality reduction techniques. It is an unsupervised learning model that captures the maximal variability of an input data using a lower dimensional space. PCA has attracted research in many different scientific fields for the last three decades. It also forms the cornerstone for developing and understanding many AI techniques, specifically in the area of Neural networks. The main task of PCA is as follows: Given $n$ data samples $\left\{ x_i\right\}_{i=1}^{n}$ where each sample lies in $\mathbb{R}^{d}$, PCA finds an orthogonal low dimensional bases where each sample can be expressed as weighted sum of these bases vectors with minimal squared error~\cite{jolliffe2002principal}. These bases vectors correspond to the eigenvectors of the covariance matrix $C=\frac{1}{n-1}\sum_{i=1}^{n} x_i x_i^T$ which can be obtained by solving the eigenvalue problem \begin{equation}
\left( C-\lambda I \right)v=0; v^T v=1,
\end {equation}
where $v \in \mathbb{R}^{d}$ is the eigenvector and $\lambda \in \mathbb{R}$ is its corresponding eigenvalue. These eigenvectors are sorted in descending order based on their eigenvalues. In practice, only a small number of eigenvectors, $k\ll n$, are chosen  to form the eigenspace.  
The main downside of PCA is that computing the eigenvectors in deterministic manner using the covariance matrix is computationally expensive, specifically when the dimensionality of samples $d$ is very large. 

Many algorithms have been developed to find the optimal eigenvectors with lower space and time complexity. We can classify each of these algorithms into two main categories: offline and online algorithms. Offline techniques compute the optimal eigenvector using a number of iterations where at each iteration a single pass over the entire dataset is performed. While using these algorithms requires the presence of all samples, they usually provide excellent convergence after very few iterations. Online approaches (also referred to as memory-limited or streaming approaches), on the other hand, aim to provide an acceptable convergence after only a single pass over the entire dataset. Thus, they are more appropriate when dealing with streaming data scenarios or when data samples are very large to fit into the memory space. While such family of algorithms considers a very practical scenario, most of these algorithms cannot be applied efficiently in these scenarios for two main reasons. Firstly, the memory limitation may significantly affect their convergence. Secondly, many of these algorithms require pre-defined parameters (such as the eigengap). These parameters require additional pre-processing pass over the data which means violating the online condition. 

In this paper, we consider the online PCA scenario. We propose an acceleration technique for such family of algorithms. Our study focuses on the empirical evaluation of such algorithms using the spiked covariance model. We will show how our acceleration scheme approaches the optimal performance without any predefined parameters or pre-processing steps. Furthermore, we evaluate our scheme using real-world time-varying scenarios.
 
\subsection*{Our Contribution}
\begin{itemize}
\item We propose an acceleration scheme for memory-limited streaming PCA that does not require any pre-defined parameters (such as the eigengap).
\item We analyze convergence guarantee in practical context using the spiked covariance model and show how our scheme achieves fast convergence even in cases where the original algorithms do not converge.
\item We employ our scheme in a practical use case of finding the key modes of motion in Molecular Dynamics simulations.
\end{itemize}

\section{Background and Literature Review}
	
	In the literature, there are two main directions that PCA research has taken. The first is that concerning applications which employ PCA for solving real-world problems and the second is that in the direction of PCA-optimization which is concerned with the optimization of the computational complexity of PCA. The link between the two directions is not clear since most studies in the application direction assume a pre-computed eigenspace and focus mainly on the distribution of test data in that eigenspace. On the other hand, in the optimization direction, the target use-case is not obvious. In addition, most of the optimization-direction algorithms are of a stochastic nature and are usually tested on rather simple datasets or data where a global eigenspase can be easily derived. In such a case, one can always consider a pre-computed eigenspace no matter what computational complexity is required for finding it. In fact, many online datasets provide a list of the most significant eigenvectors of the studied samples.
	
	With regard to the applications research, the use of PCA has been well reported in the fields such as Computer Vision and Computer Graphics. For instance, in facial recognition, Kirby and Sirovich \cite{kirby1990application}
	proposed PCA as a holistic representation
	of the human face in 2D images by extracting few orthogonal dimensions
	which form the face-space and were called eigenfaces \cite{turk1991eigenfaces}. Gong et al. \cite{gong1996investigation} were the first to find the relationship between the distribution of samples in the eigenspace, which were called manifolds, and the actual pose in an image of a human face. 
	The use of PCA was extended using Reproducing Kernel Hilbert Spaces which non-linearly map
	the face-space to a much higher dimensional space (Hilbert space)
	~\cite{yang2002kernel}.  Knittel and Paris~\cite{knittel09pcaseeding} employed a PCA-based technique to find initial seeds for vector quantization in image compression.
		
	In the PCA-optimization research, the power iteration remains one of the most popular offline techniques for finding the top $k$ eigenvectors \cite{golub2012matrix}. It has an exponential convergence rate when the eigengap $\Delta=\lambda_{1}-\lambda_{2}$ (the difference between eigenvalues of the first and second eigenvectors) is large enough. However, when the eigengap is small, this method converges very slowly. Shamir solves this problem by applying the stochastic Variance Reduction technique~\cite{johnson2013accelerating} in conjunction with the power iterations (VR-PCA) achieving a much faster convergence rates even in cases where the eigengap is too small~\cite{shamir2015stochastic}. Recently,~\cite{xu2018accelerated} proposed directly adding the momentum term to both the power iteration and VR-PCA for better convergence rates. In terms of the online PCA algorithms, the update schemes proposed by Krasulina \cite{krasulina1969method} and Oja \cite{oja1982simplified,oje1983subspace} are amongst the most popular online PCA techniques. The speed of convergence of these techniques depends mainly on the learning rate. Such a choice of learning rate is a matter of ongoing research. Balsubramani et al. \cite{balsubramani2013fast} found that optimal convergence rate of Oja's rule approaches $\mathcal{O}(1/t)$ when scaling the learning rate based on the eigengap. However, a prior knowlege of the eigengap may be not possible when dealing with online scenarios. \cite{allen2017first} suggested eigengap-free learning rates for such cases. However, no experimental results were provided for their technique. Mitiagkas et al. proposed an online non-parametric PCA algorithm for streaming data based on a simplified version of the power iterations which we refer to as the Block Power Method~\cite{mitliagkas2013memory}. The main idea is that at each time-step, the power update step is applied on a subset (block) of the dataset revealed at time $t$. The downside of this technique is that the block size depends on the number of dimensions per time-step. Hence, choosing very small block size will give very poor convergence. Li et al. found that changing the block size among different iterations may significantly enhance the quality performance~\cite{li2016rivalry}.  More recently, Xu et al. suggested adding the momentum term to accelerate convergence of state of the art PCA complexity optimization algorithms~\cite{xu2018accelerated}. Particularly, they applied the momentum model on Oja's method, block power, batch power iterations and VR-PCA. Their results showed that the momentum model significantly enhances offline techniques while achieving a limited improvement when applied on the the limited memory, online approaches.
	
\subsection{Problem Formulation}
Given a time-varying dataset $X=\left[x_{t_{1}},\,x_{t_{2}},\ldots,\,x_{t_{n}}\right]\in\mathbb{R}^{d\times n}$ where $d$ is the total number of dimensions (attributes), our goal is to find the top $k$ eigenvectors from a single pass of the data without any prior knowledge about the distibution of input samples. This is particularly important in cases where a single sample may be too large to fit into the memory space. Time-varying data are more challenging from many perspectives. Firstly they usually do not satisfy the i.i.d. assumption which most online PCA techniques require. In other words, consecutive frames are not completely independent. Secondly, in case of streaming data scenarios, it is very hard to estimate key properties of the dataset such as the general distribution and eigengap between first two eigenvectors. 	
	
	\subsection{Main result}
	Figure~\ref{fig:example} shows an overview example demonstrating the main goal of this paper. One can see at a glance that the block power when using very small block size does not converge at all while Oja's rule, when using an arbitrary learning rate, converges very slowly towards the first significant eigenvector. One solution is to use the optimal learning rate of Oja's rule according to~\cite{balsubramani2013fast} which significantly improves the convergence rate. However, as mentioned earlier, finding such optimal learning rate violates the online learning condition as it requires an additional preprocessing data pass to compute the eigengap $\Delta$. Our acceleration technique remedies this problem and requires neither prior knowledge of the data samples nor restriction on the block size. With a single pass on the generated data, our scheme converges to the first significant eigenvector reaching $0.9999$ accuracy when accelerating the block power and $0.999$ when accelerating Oja's method with learning rate of $\eta_t=1/t$. Here, we define accuracy as $accuracy=1-convergence$ and $convergence=1-\frac{\left\Vert X^{T}W_{t}\right\Vert _{F}^{2}}{\left\Vert X^{T}V^{*}\right\Vert _{F}^{2}}$ where $W_t$ is the eigenvector estimate at time $t$ and $V^*$ is the optimal eigenvector. Furthermore, we will see later in this study how our technique works efficiently with real-world time-varying data.


\section{Our Accelerations Scheme}
\begin{figure}[h]
		\centering
			\includegraphics[width=0.9\linewidth]{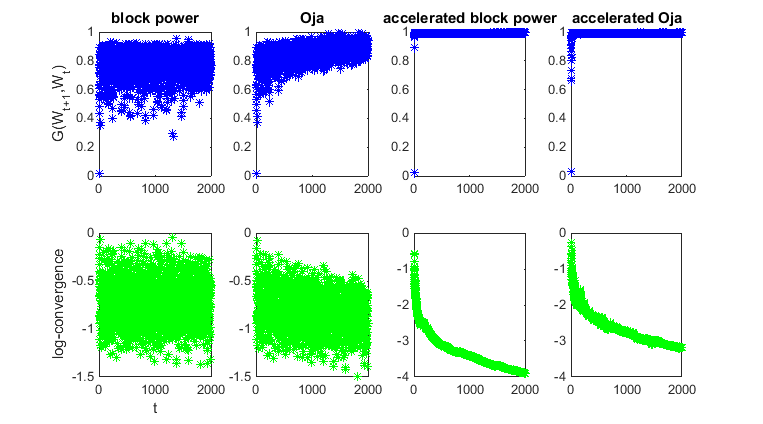}
		\caption{\label{fig:acceleration}
			Performance analysis of each online method in terms of the objective function $G\left(W_{t+1},W_{t}\right)$ (Top) and the log-convergence defined as $\log_{10}\left(1-\frac{\left\Vert X^{T}W_{t}\right\Vert _{F}^{2}}{\left\Vert X^{T}V^{*}\right\Vert _{F}^{2}}\right)$ (Bottom) when computing the first eigenvector of the spiked covariance model.}
	\end{figure}

In this section, we will explain our acceleration scheme. Our scheme
is based on a very simple yet efficient idea. Assume that the acceleration
function $g\left(W_{t+1},W_{t},t\right)$ takes as an input the updated
eigenvectors at times $t$ and $t+1$ and the number of current update,
the main objective is to maximize the following function
\[
G\left(W_{t+1},W_{t}\right)=W_{t+1}^{T}W_{t}W_{t}^{T}W_{t+1}
\]
where $W_t$ and $W_{t+1}$ are normalized and lie in $\mathbb{R}^{d}$ space. This maximization satisfies an important condition of convergence for all online PCA
algorithms. It follows naturally by directly applying the gradient
ascend rule on the objective function one gets 
\begin{align*}
g\left(W_{t+1},W_{t},t\right)&=W_{t+1}+\frac{\alpha_{t}}{2}\frac{\partial}{\partial W_{t+1}}G\left(W_{t+1},W_{t}\right)\\
&=W_{t+1}+\alpha_{t}W_{t}W_{t}^{T}W_{t+1}\\
&=\left(I+\alpha_{t}W_{t}W_{t}^{T}\right)W_{t+1}
\end{align*}
where $\alpha_{t}$ is the learning rate for which we will justify its
values later in this study. This leads to the general update rule
\begin{equation}
W_{t+1}=\frac{\tilde{W}_{t+1}+\alpha_{t}W_{t}W_{t}^{T}\tilde{W}_{t+1}}{\left\Vert \tilde{W}_{t+1}+\alpha_{t}W_{t}W_{t}^{T}\tilde{W}_{t+1}\right\Vert },
\end{equation}
where $\tilde{W}_{t+1}=f\left(W_{t},\:\tilde{X}_{t}\right)$ is the eigenvector estimation based on the update method $f$ that we wish to accelerate. Here, $\tilde{X}_t=\left\{x_i \right\}_{i=Bt+1}^{B(t+1)} $ is a subset of the input dataset revealed at time $t$ with cardinality $B$ and $W_t \in \mathbb{R}^{d}$ is the recent updated eigenvector. We call this subset a block of samples of size $B$. For instance, in case of Oja's rule $f\left(W_t,\: \tilde{X}_t \right)=W_t+\eta\left(t\right) \tilde{X}_{t}\tilde{X}_t^{T} W_t/B$ and in case of block power $f\left(W_t,\: \tilde{X}_t \right)=\tilde{X}_{t}\tilde{X}_t^{T}W_{t}/B$. Note that here we consider a block variant of Oja's method, where at each iteration we update the eigenvector based on a block of recent time-steps instead of using only the most recent one. Our scheme overcomes the memory limitations by emphasizing
on the shared information acquired from previous eigenvectors instead
of depending heavily on the recent time-steps as the case in Oja and
block power which may not reflect the temporal changes throughout
all observations. This will give more robustness against new samples
corresponding to outliers. In order to show how our acceleration rapidly
optimizes $G\left(W_{t+1},W_{t}\right)$, we will compare the values
of the function for the block power and Oja's methods before and after
applying our acceleration scheme as shown in Figure~\ref{fig:acceleration}. It's very clear
that using our scheme $G\left(W_{t+1},W_{t}\right)$ reaches the
optimal value of 1 after very few iterations with much better and more stable performance in terms of convergence rates. Our scheme can be easily generalized to extract multiple eigenvectors ($k>1$ case) by defining $W_t \in \mathbb{R}^{d \times k}$ and replacing the normalization process by Gram-Schmidt orthogonalization process.

\section{Experimental Design}

We test our method on synthetic datasets generated using the spiked covariance model based on~\cite{mitliagkas2013memory} where each time-step is drawn from the following generative model \[
x_{i}=Az_{i}+\sigma N_{i},
\]
where $A\in\left[-1,1\right]^{d\times k}$ is a fixed matrix, $z_{i}\in\mathbb{R}^{k}$
is a random weight vector based on standard normal distribution and $\sigma N_{i}\in\mathbb{R}^{d}$
is a Gaussian noise vector of standard deviation $\sigma$. The task
is to restore the component matrix $A$ from the noisy samples $x_i$. We test the online techniques for the following settings $k=1$ and $10$, $d=100$ and $1,000$ and $\sigma =0.5$ and $1$. We apply our scheme for accelerating the block power and Oja's algorithms. For the block power, the block size was set to $B=5$ in order to show how our model converges even when using very small block size. For Oja's rule, the learning rate was set according to $\eta\left(t\right)=\frac{ac_{t}+1}{t}$ where $c_{t}$ is a random variable of uniform
distribution and $a=2$. For our scheme, we set the learning rate to $\alpha_t=1/\eta(t)$. Hence, the update rule of our scheme becomes \[
W_{t+1}=\frac{\tilde{W}_{t+1}+\frac{t}{ac_{t}+1}W_tW_{t}^{T}\tilde{W}_{t+1}}{\left\Vert \tilde{W}_{t+1}+\frac{t}{ac_{t}+1}W_{t}W_{t}^{T}\tilde{W}_{t+1}\right\Vert }.
\] This allows for better performance analysis by assuming our acceleration scheme to be a stochastic process. All methods were initialized using same random vector from unit sphere. At each time-step, convergence to the optimal eigenvectors is evaluated in terms of the log-convergence as follows  \[
{log-convergence}\left(V^{*},W_{t}\right)=\log_{10}\left(1-\frac{\left\Vert X^{T}W_{t}\right\Vert _{F}^{2}}{\left\Vert X^{T}V^{*}\right\Vert _{F}^{2}}\right)
\] where $\left\Vert.\right\Vert_{F}^{2}$ is the squared Frobenius norm and $ V^* \in \mathbb{R}^{d \times k}$ are the optimal $k$ eigenvectors computed using batch PCA.
\subsection{Results} 
\begin{figure}[t]
		\centering 
		
		\begin{minipage}{1\textwidth}
			\centering
			\includegraphics[width=1.0\linewidth]{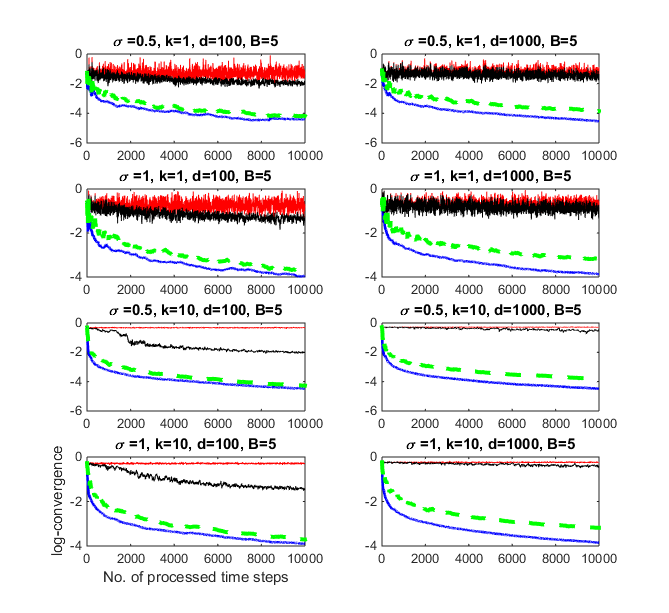}
			
		\end{minipage}
		\begin{minipage}{0.3\textwidth}
			\centering
			\includegraphics[width=1.0\linewidth]{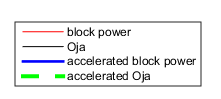}
		\end{minipage}

		\caption{\label{fig:Results}
			Convergence of each technique to the first 10 eigenvectors of the spiked covariance model for different settings.}
	\end{figure}

Fig.\ref{fig:Results} shows the performance of each technique on the spiked model for different experimental settings. It is very clear that our scheme gives fastest convergence rates in all settings reaching $0.99$ accuracy before processing $10\%$ of the data. After processing all time-steps, the convergence of our scheme becomes $<10^{-3}$. In general, the accelerated block power performs better than accelerated Oja. The block power (without acceleration) on the other hand was not converging in all experiments because of the very small block size. Oja's method converges very slowly specially in the case of $d=1,000$ with very clear oscillating behaviour when extracting a single eigenvector ($k=1$ case).

\subsection{Online PCA for Analyzing Molecular Dynamics}
 
\begin{figure*}[h]
		\centering
		\begin{minipage}{0.45\textwidth}
			\centering
			\includegraphics[width=1.0\linewidth]{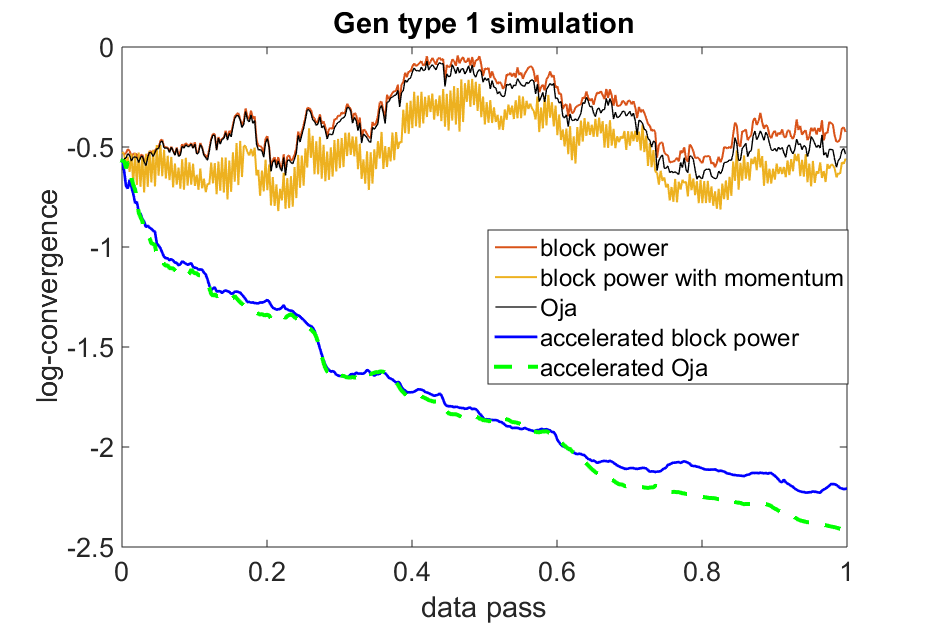}
			
		\end{minipage}
		\begin{minipage}{0.45\textwidth}
			\centering
			\includegraphics[width=1.0\linewidth]{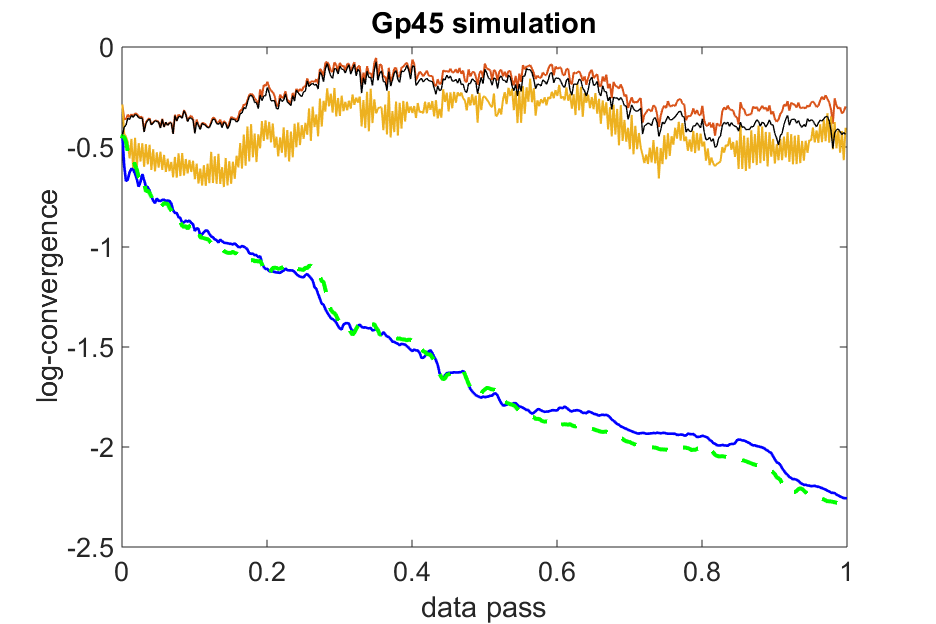}
		\end{minipage}\\
		
        \begin{minipage}{0.45\textwidth}
			\centering
			\includegraphics[width=1.0\linewidth]{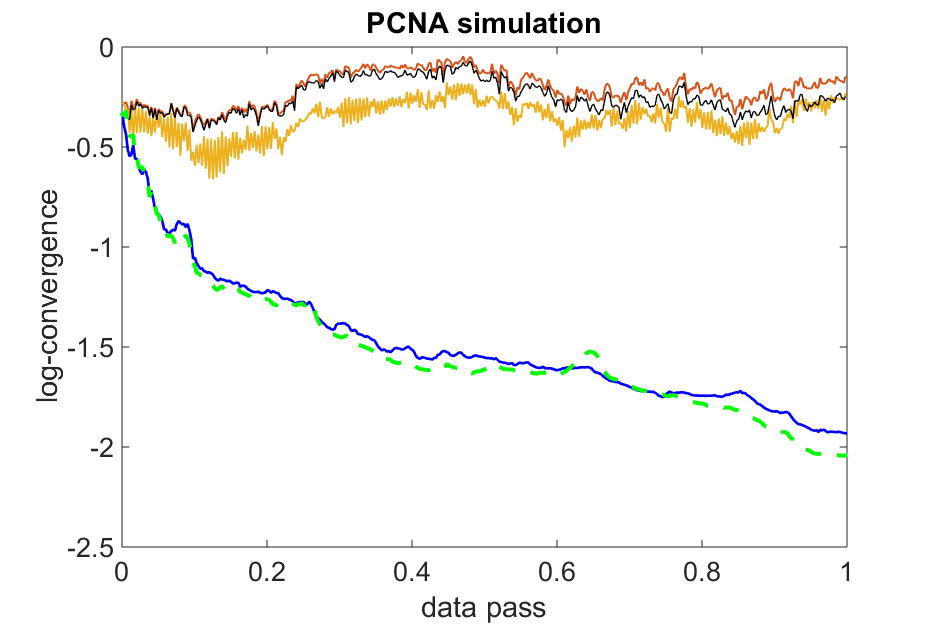}
			
		\end{minipage}

		\caption{\label{fig:Results_time_varying}
			Convergence of each technique to the first 20 eigenvectors of three Molecular Dynamics simulations.}
	\end{figure*}

In this section, we will study the performance of online PCA methods for analyzing trajectories in Molecular Dynamics simulations where atoms interact with each other in complex spring-like motion. The use of PCA for analyzing MD trajectories, specifically dynamics in proteins, has become a widespread since the work done by~\cite{amadei1993essential}. It has been reported that standard PCA converges to the actual normal modes in cases where atoms have harmonic motion.  However, the major problem in MD simulation is that trajectories are usually represented using the Cartesian coordinates of each atom at time $t$. Hence the covariance matrix becomes of size $3N \times 3N$ where $N$ is the number of atoms. When the number of atoms is large (which is the case in most simulations) the covariance matrix becomes too large to fit in the memory space. To the best of our knowledge, the performance of online PCA for such systems has not been reported. We apply online PCA techniques on three MD simulations from~\cite{oakley2016dynamics}. Table~\ref{tab:summary_dataset} shows a summary of the datasets. In all experiments, we set the block size to $B=5$. Figure~\ref{fig:Results_time_varying} shows convergence to the first 20 eigenvectors using each technique. In all experiments, both Oja's method and block power do not converge. We can also note that adding the momentum term to the block power gives a very limited improvement. On the other hand, after applying our acceleration technique, we get an accuracy of $0.99$ after single data pass.

\begin{table}[h]
\caption{\label{tab:summary_dataset}
            Summary of each MD dataset/simulation.}
\centering{}%
\begin{tabular}{|c|c|c|c|}
\hline 
\textbf{\scriptsize{}dataset/simulation name} & \textbf{\scriptsize{}No. of dimensions} & \textbf{\scriptsize{}No. of time-steps}\tabularnewline
\hline 
\textbf{\scriptsize{}PCNA} & {\scriptsize{}36,675} & {\scriptsize{} 2,000}\tabularnewline
\hline 
\textbf{\scriptsize{}gen-type 1} & {\scriptsize{}34,248} & {\scriptsize{}2,000}\tabularnewline
\hline 
\textbf{\scriptsize{}Gp45 bindings} & {\scriptsize{}31,806} & {\scriptsize{}2,000}\tabularnewline
\hline
\end{tabular}
\end{table}

\section{Discussion}
In this paper, we investigated the problem of finding the first $k$ eigenvectors of any dataset in a single pass. We found that state-of-the-art online PCA methods fail to solve this problem in real world scenarios. We proposed an acceleration scheme for such family of algorithms. We focused on the empirical evaluation of our scheme using the spiked covariance model. The convergence rate after applying our scheme is much faster than the original online methods. We further tested our scheme on real-world time-varying datasets of Molecular Dynamics simulations. While online methods fail to converge in such scenarios, our scheme recovered the top eigenvectors with accuracy of $0.99$ after a single data pass. In terms of our future work, we would like to have more theoretical analysis on our scheme.

\section{Acknowledgements}
This research has been conducted with the financial support of Science Foundation Ireland (SFI) under Grant Number 13/IA/1895.

\bibliographystyle{splncs}
\bibliography{egbib}

\begin{thebibliography}{10}

\bibitem{jolliffe2002principal}
Jolliffe, I.:
\newblock Principal component analysis.
\newblock Wiley Online Library (2002)

\bibitem{kirby1990application}
Kirby, M., Sirovich, L.:
\newblock Application of the karhunen-loeve procedure for the characterization
  of human faces.
\newblock Pattern Analysis and Machine Intelligence, IEEE Transactions on
  \textbf{12} (1990)  103--108

\bibitem{turk1991eigenfaces}
Turk, M., Pentland, A.:
\newblock Eigenfaces for recognition.
\newblock Journal of cognitive neuroscience \textbf{3} (1991)  71--86

\bibitem{gong1996investigation}
Gong, S., McKenna, S., Collins, J.J.:
\newblock An investigation into face pose distributions.
\newblock In: Automatic Face and Gesture Recognition, 1996., Proceedings of the
  Second International Conference on, IEEE (1996)  265--270

\bibitem{yang2002kernel}
Yang, M.H.:
\newblock Kernel eigenfaces vs. kernel fisherfaces: Face recognition using
  kernel methods.
\newblock In: Proceedings of Fifth IEEE International Conference on Automatic
  Face Gesture Recognition. (2002)  215--220

\bibitem{knittel09pcaseeding}
Knittel, G., Parys, R.:
\newblock {PCA}-based seeding for improved vector quantization.
\newblock In: Proceedings of the First International Conference on Computer
  Imaging Theory and Applications (VISIGRAPP 2009). (2009)  96--99

\bibitem{golub2012matrix}
Golub, G.H., Van~Loan, C.F.:
\newblock Matrix computations. Volume~3.
\newblock JHU Press (2012)

\bibitem{johnson2013accelerating}
Johnson, R., Zhang, T.:
\newblock Accelerating stochastic gradient descent using predictive variance
  reduction.
\newblock In: Advances in neural information processing systems. (2013)
  315--323

\bibitem{shamir2015stochastic}
Shamir, O.:
\newblock A stochastic pca and svd algorithm with an exponential convergence
  rate.
\newblock In: ICML. (2015)  144--152

\bibitem{xu2018accelerated}
Xu, P., He, B., De~Sa, C., Mitliagkas, I., Re, C.:
\newblock Accelerated stochastic power iteration.
\newblock In: International Conference on Artificial Intelligence and
  Statistics. (2018)  58--67

\bibitem{krasulina1969method}
Krasulina, T.:
\newblock A method of stochastic approximation for the determination of the
  least eigenvalue of a symmetric matrix.
\newblock Zhurnal Vychislitel'noi Matematiki i Matematicheskoi Fiziki
  \textbf{9} (1969)  1383--1387

\bibitem{oja1982simplified}
Oja, E.:
\newblock Simplified neuron model as a principal component analyzer.
\newblock Journal of mathematical biology \textbf{15} (1982)  267--273

\bibitem{oje1983subspace}
OJE, E.:
\newblock Subspace methods of pattern recognition.
\newblock In: Pattern recognition and image processing series. Volume~6.,
  Research Studies Press (1983)

\bibitem{balsubramani2013fast}
Balsubramani, A., Dasgupta, S., Freund, Y.:
\newblock The fast convergence of incremental {PCA}.
\newblock In: Advances in Neural Information Processing Systems. (2013)
  3174--3182

\bibitem{allen2017first}
Allen-Zhu, Z., Li, Y.:
\newblock First efficient convergence for streaming k-pca: a global, gap-free,
  and near-optimal rate.
\newblock In: Foundations of Computer Science (FOCS), 2017 IEEE 58th Annual
  Symposium on, IEEE (2017)  487--492

\bibitem{mitliagkas2013memory}
Mitliagkas, I., Caramanis, C., Jain, P.:
\newblock Memory limited, streaming pca.
\newblock In: Advances in Neural Information Processing Systems. (2013)
  2886--2894

\bibitem{li2016rivalry}
Li, C.L., Lin, H.T., Lu, C.J.:
\newblock Rivalry of two families of algorithms for memory-restricted streaming
  pca.
\newblock In: Artificial Intelligence and Statistics. (2016)  473--481

\bibitem{amadei1993essential}
Amadei, A., Linssen, A., Berendsen, H.J.:
\newblock Essential dynamics of proteins.
\newblock Proteins: Structure, Function, and Bioinformatics \textbf{17} (1993)
  412--425

\bibitem{oakley2016dynamics}
Oakley, A.J.:
\newblock Dynamics of open dna sliding clamps.
\newblock PloS one \textbf{11} (2016)  e0154899

\end{thebibliography}

\end{document}